\documentclass[letterpaper, 10 pt, conference]{ieeeconf}  

\IEEEoverridecommandlockouts                              
\usepackage{hyperref}
\usepackage{graphicx}

\usepackage[utf8]{inputenc}
\usepackage[T1]{fontenc}
\usepackage{flushend}
\begin{document}


\title{\LARGE \bf Safe Spot: Perceived safety of dominant and submissive appearances of quadruped robots in human-robot interactions}

\author{Nanami Hashimoto$^{1}$, Emma Hagens $^{2}$, Arkady Zgonnikov$^{3}$ and Maria Luce Lupetti$^{4}$
\thanks{$^{1}$Department of Cognitive Robotics, Delft University of Technology, Delft, The Netherlands
        {\tt\small N.Hashimoto@student.tudelft.nl}}%
\thanks{$^{2}$Department of Human Centred Design, Delft University of Technology, Delft, The Netherlands
        {\tt\small Emmahagens98@gmail.com}}%
\thanks{$^{3}$Department of Cognitive Robotics, Delft University of Technology, Delft, The Netherlands
        {\tt\small A.Zgonnikov@tudelft.nl}}%
\thanks{$^{4}$Department of Architecture and Design, Politecnico di Torino, Turin, Italy
        {\tt\small maria.lupetti@polito.it}}%
}%
\maketitle

\begin{abstract}
Unprecedented possibilities of quadruped robots have driven much research on the technical aspects of these robots. However, the social perception and acceptability of quadruped robots so far remain poorly understood. This work investigates whether the way we design quadruped robots' behaviors can affect people’s perception of safety in interactions with these robots. We designed and tested a dominant and submissive personality for the quadruped robot (Boston Dynamics Spot). These were tested in two different walking scenarios (head-on and crossing interactions) in a 2x2 within-subjects study. We collected both behavioral data and subjective reports on participants' perception of the interaction. The results highlight that participants perceived the submissive robot as safer compared to the dominant one. The behavioral dynamics of interactions did not change depending on the robot's appearance. Participants' previous in-person experience with the robot was associated with lower subjective safety ratings but did not correlate with the interaction dynamics. Our findings have implications for the design of quadruped robots and contribute to the body of knowledge on the social perception of non-humanoid robots. We call for a stronger standing of felt experiences in human-robot interaction research.
\end{abstract}

\section{INTRODUCTION}
Driven by a vision of unprecedented possibilities \cite{raibert2008bigdog}, quadruped robots have been received increasing attention both from the industry and the academic community \cite{ferreira2022survey}.
Compared to their wheeled counterparts, quadruped robots can better handle environments with multiple obstacles and uneven terrains \cite{ferreira2022survey, biswal2021development}, which affords practical advantages to several industries now developing applications in various domains, including entertainment, goods delivery, hazardous areas inspection, and law enforcement \cite{biswal2021development, George2021}.

Along with the excitement, however, also come concerns for the societal implications that these types of robots may have. In particular, quadruped robots may be used in support of controversial activities, such as surveillance and law enforcement, or even as autonomous weapon systems \cite{moses2021see}. A multitude of civil uses are being explored, yet the public perception of these robots remains strongly influenced by the doom scenarios \cite{moses2021see} evoked by the applications being tested in the police and military domains \cite{biswal2021development, moses2021see}, which regularly receive extensive media coverage. Companies like Boston Dynamics have put explicit effort into discouraging these negative views, such as by stating that their platforms are intended \emph{“to help, not to harm”} \cite{Boston2023} and thus cannot be weaponized (stated also in their Terms and Conditions of Sale \cite{Ackerman2021}). 

Media and activists, however, continue to contest these actions as mere "good intentions" and underscore the possible risks of quadruped robots. For instance, the art collective MSCHF created \emph{Spot’s Rampage}, a performance artwork \cite{Hambling2021} in which the robot was armed with a paintball gun and then connected to the internet for remote players to control it. Through this, they illuminate how easy it could be to misuse these platforms.

Academic research, on the contrary, has hardly engaged with social aspects of quadruped robots. The literature on these robots remains mostly focused on technical matters, such as locomotion, localization, mapping \cite{ferreira2022survey}, and environment perception \cite{meng2016review}. Only a few studies have explored social perception and acceptability of quadruped robots so far, which leaves these aspects poorly understood and unattended \cite{hauser2023s}.

\begin{figure}[!t]
  \includegraphics[width=\linewidth]{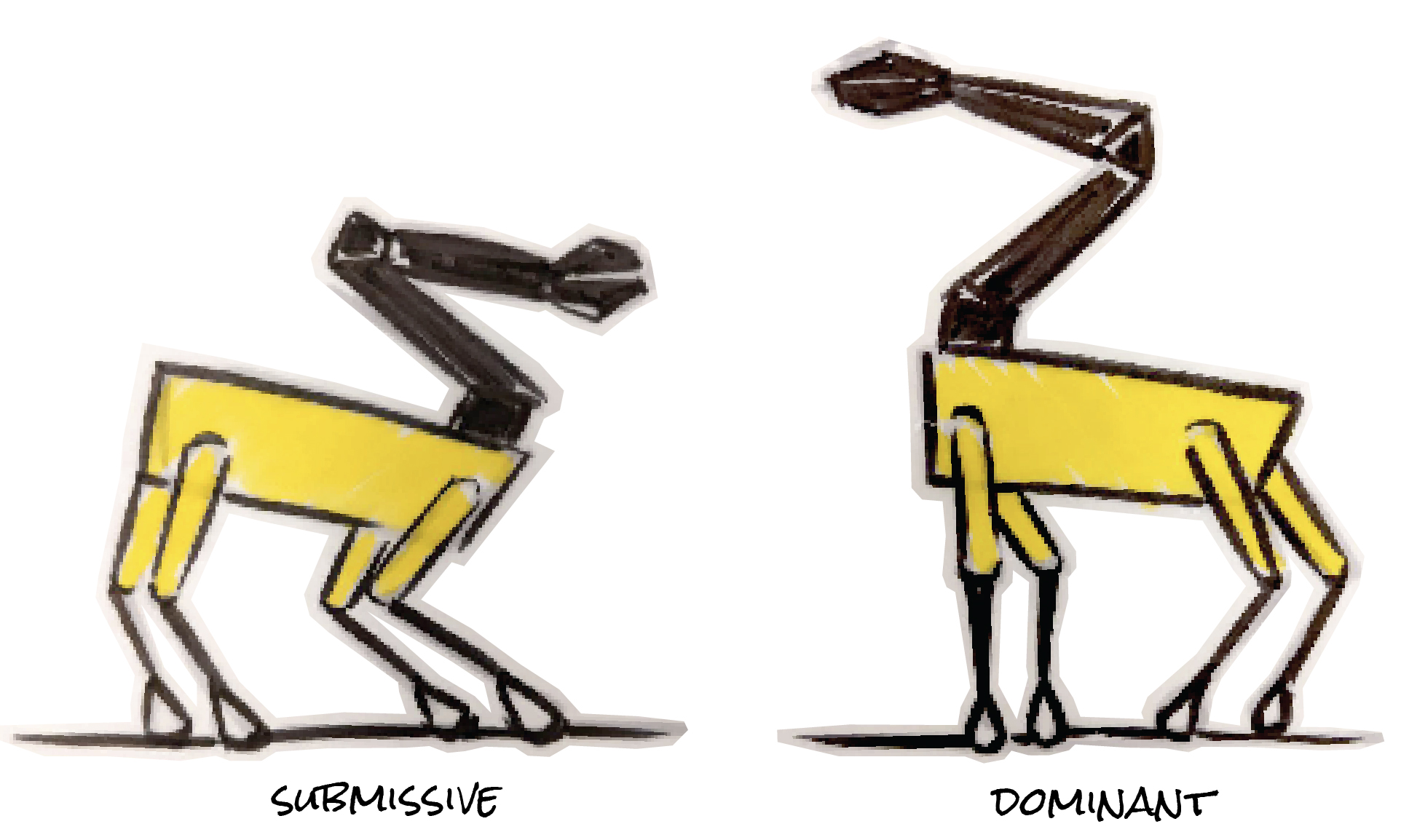}
  \caption{Submissive and dominant personalities for quadruped robots.}
  \label{teaser}
\end{figure}

This work aims to contribute to filling this gap by investigating whether the design of robotic behaviors can affect people’s perception of quadruped robots. In particular, we compare accidental encounters with a quadruped robot displaying two alternative personalities--dominant vs submissive (Figure~\ref{teaser}) -- and assess these in terms of perceived safety. As studies found before, perceived safety --- the user’s perception of danger and comfort when interacting with a robot --- is a determining factor in achieving robot acceptability in human-inhabited environments \cite{bartneck2009measurement}. Through our findings, we complement existing knowledge on media representation and public perceptions of quadruped robots (e.g., through social media analysis \cite{moses2021see}), with a distinctively empirical view into the felt experience of people when interacting with these robots.

\section{SOCIAL PERCEPTION OF QUADRUPED ROBOTS}
The more quadruped robots are adopted in public spaces, the more they are required to understand and adapt not only to the environment and users but also and foremost to accidentally co-present people \cite{moesgaard2022incidental}. Literature on the social perception of quadruped robots, however, is limited. For instance, in their exploration of quadruped robots for social distancing (in COVID emergency scenarios), Chen et al. \cite{chen2021autonomous} report a generally positive response toward their envisioned use of quadruped robots, with only a few exceptions. The results, however, are presented as brief insights and do not provide further articulation of how the robot was perceived and approached by participants. To our knowledge, the only extensive study that looked at the social perception of quadruped robots is the work by Moses and Ford \cite{moses2021see}. Through a sentiment analysis of data extracted from Twitter posts, the authors observed how corporate advertisement videos and other media portrayals had a direct effect on the sentiment of the public towards quadruped robots, varying constantly from positive to negative views. The study provides insights into how these robotic platforms are both perceived as inspiring and exciting new technological marvels, as well as associated with "killing robots" and as such, feared by the public. While rich and insightful, also this study provides only a partial view of the complex space of social perception of quadruped robots as it focuses on media portrayals and online discourse. 

Empirical evidence about the felt experience of people in the presence of quadruped robots is still mostly lacking.
These, however, are implicitly present in the adjacent discourse on the social navigation of quadruped robots, as such navigation inherently requires understanding what makes robots socially competent and acceptable for people.

\subsection{Social navigation of quadruped robots}
\emph{Social robot navigation} is a key challenge when robots operate in civil environments, which requires a rich understanding of the technical and behavioral, as well as social issues at play \cite{mavrogiannis2023core, rios2015proxemics}. 
Together with ensuring efficient path planning and physical safety, robots are required to be aware and responsive to social conventions \cite{rios2015proxemics}, such as proxemics, and socially relevant needs, i.e. effective communication of intentions \cite{mavrogiannis2023core}. Thereafter, social robot navigation is an ideal site for studying the social perception of robots in action, where people might be "forced" to deal with robots unexpectedly and with physical constraints. 

The topic has received great attention from the human-robot interaction (HRI) community in the last two decades \cite{mavrogiannis2023core}. Few studies, however, focus on the social navigation of quadruped robots specifically and these few address only partial aspects of the problem. For instance, Karnan et al. \cite{karnan2022socially} curated and validated a dataset of demonstrations (i.e., lidar scans, joystick commands, odometry, camera visuals, and 6D inertial) for socially compliant navigation in the wild to be used for imitation learning. Yang et al. \cite{yang2022online} developed efficient and socially-aware motion planning models by focusing on maintaining high social distance and ensuring smooth velocity transitions. Relatedly, Katyal et al. \cite{katyal2022learning}, together with socially desirable distancing, explored group dynamics to inform and improve the path planning of quadruped robots.

While important, these works maintain a focus on the capabilities of the robot (i.e., improve path planning) while matters of human perception remain underexplored. Only a few authors have looked at how quadruped robots specifically are perceived by people. Hauser et al. \cite{hauser2023s} found that using explicit visual signs of canine control (i.e. a human holding the robot on a leash) can positively influence participants’ experience during incidental encounters with the robot. Xu et al. \cite{xu2023understanding}, instead, learned that the orientation and gaze of the quadruped robot can affect the personal distance that people prefer to keep when co-present. Specifically, their participants preferred keeping a greater distance when the robot is static, as they found it more difficult to understand the robot intentions and worried about sudden movements. 
Both Hauser et al. \cite{hauser2023s} and Xu et al. \cite{xu2023understanding}, however, left unattended the aspects of perceived safety and attitude towards quadruped robots that have emerged prominently in the societal discourse around these platforms. Last, Sanoubari et al. \cite{sanoubari2022message} investigated and confirmed that improving communication with quadruped robots by enhancing their expressivity leads to higher perceived safety. In particular, the authors designed and evaluated six expressions (combinations of visual and gestural cues), such as examining, noticing, greeting, and more. Despite the positive results, however, the study focused on intentional aspects of communication between the robot and people (e.g., communication of intention and greeting). Insights into people's perception of quadruped robots, when no direct interaction is planned, are still lacking.

\subsection{Perceived safety as facilitator of social acceptability}
Understanding how safe people feel in the presence of quadruped robots is key to understanding their social perception more broadly. Perceived safety is described as “the user’s perception of the level of danger when interacting with a robot, and the user’s level of comfort during the interaction” \cite{bartneck2009measurement}. Compared to physical safety, which is about preventing robots from causing physical harm (e.g., via collision), this emphasizes the key role that people’s psychological perception of the interaction with robots has on their ultimate acceptability and adoption \cite{sanoubari2022message, zacharaki2020safety}. 

People may feel unsafe even when no physical risks are present \cite{akalin2023taxonomy, sanoubari2022message} because many factors affect the user experience of a robot, such as the context of use, comfort, familiarity, predictability and transparency, sense of control, and trust \cite{akalin2023taxonomy}. Already more than a decade ago, Bartneck et al. \cite{bartneck2009measurement} argued for the need to systematically and directly study perceived safety and user comfort. Despite the topic receiving increasing attention, however, the field still ``\textit{lacks a comprehensive investigation of perceived safety towards an enduring presence of robots sharing a common space with humans}'' \cite{akalin2023taxonomy}. And this knowledge gap is even more prominent within the space of quadruped robot HRI research. Some work \cite{hauser2023s, xu2023understanding} provides insights on and methods for achieving predictability and transparency (important aspects of perceived safety). Yet we still lack data on whether and how the very appearance and the related personality of quadruped robots affect perceived safety.

\section{EXPLORING THE PERCEIVED SAFETY OF SUBMISSIVE VS DOMINANT QUADRUPED ROBOTS}
This work sets out to provide insights into people’s perceived safety of quadruped robots and, more specifically, whether the design of robots' non-verbal behaviors may have an effect on this. Similarly to Sanoubari et al. \cite{sanoubari2022message}, we explore robot expressivity as a facilitating factor in perceived safety but distinctively focus on the implicit communication of personalities and related values \cite{demirbilek2003product, desmet2015emotions}, rather than practical communicative functions (e.g., intention to turn). We focus on bodily postures, which can be a strong medium for communicating robot personalities \cite{zabala2021expressing, hoffman2014designing, meerbeek2009towards} with direct effects on people’s impressions \cite{hiah2013abstract, kim2008personality} and desirable proxemics \cite{obaid2016stop}.

\begin{figure}[ht!]
  \centering
  \includegraphics[width=1\linewidth]{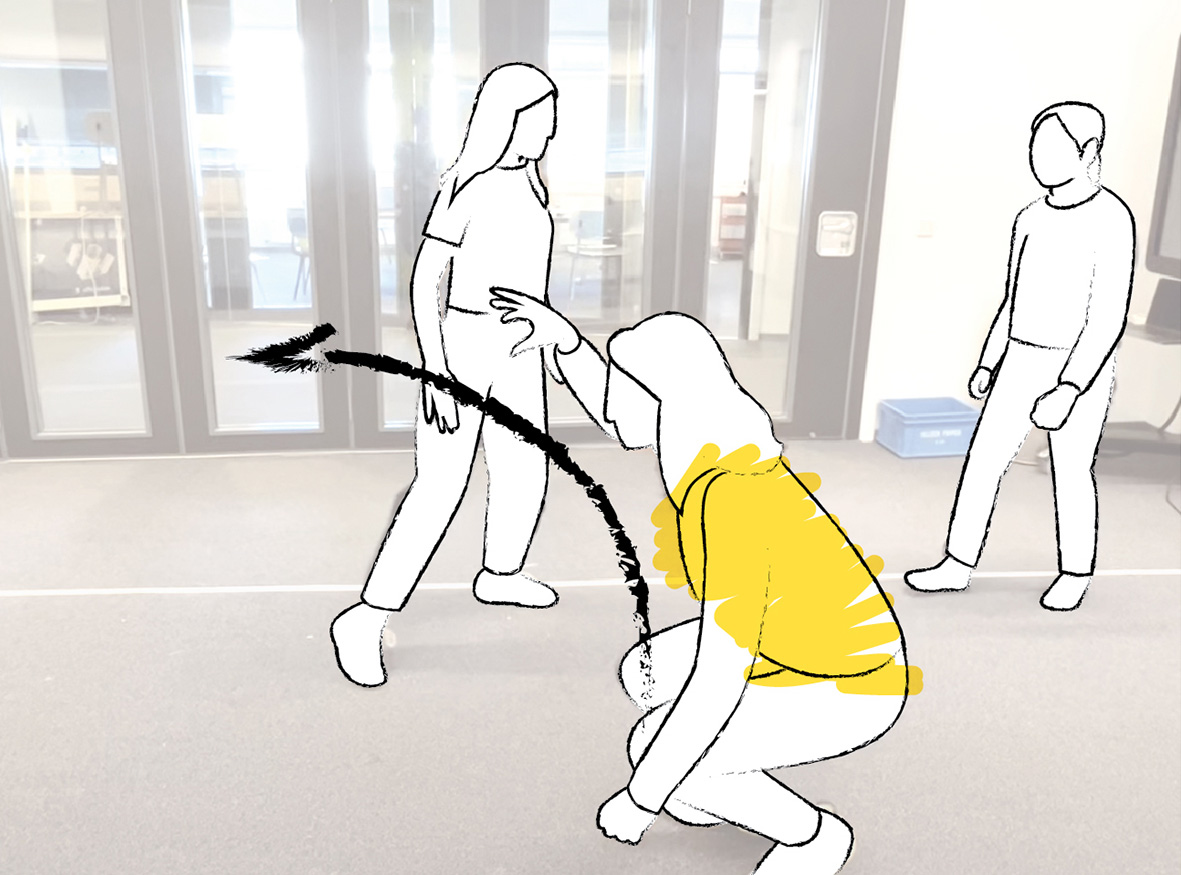}
  \caption{Bodystorming session: one researcher impersonates the robot (highlighted in yellow) and enacts a submissive personality, while two other researchers enact a nurse and a person visiting the hospital}
  \label{fig:bodystorming}
\end{figure}

Which robot personality should be preferable for achieving perceived safety and social acceptability is dependent on multiple factors including people’s value systems \cite{chatzoglou2023factors}, the perceived usefulness of the robot \cite{de2019would}, and their typology \cite{naneva2020systematic}. A robotic closet with a submissive personality was shown to be preferred to a dominant one \cite{hiah2013abstract}, while a humanoid robot was reported to be more enjoyable and positively received by people when showing an extrovert and confident personality \cite{kim2008personality}. With regards to quadruped robots, however, no research so far has explored whether and how a different personality of the robot would affect perceived safety (and thus acceptability).
Our work, then, aims to fill this gap by investigating the possibility of the robot showing a submissive personality, as opposed to the confident and dominant personality we usually witness in advertisements and popular media portraying quadruped robots \cite{moses2021see}. We ideated, prototyped, and evaluated a \textit{dominant} and a \textit{submissive} personality for a quadruped robot (Spot by Boston Dynamics) by combining a Research through Design approach \cite{zimmerman2007research} with embodied ideation and empirical evaluation.

\subsection{Embodied ideation of robot behaviors} 
We built upon a growing number of human-computer and human-robot interaction studies that employ performative methods  \cite{belling2021rhythm, marquez2016bodystorming, lee2014practicing, youaffective}, to understand felt experiences and social interactions with technologies \cite{luria2020robotic}. 

We conducted a two-hour embodied ideation session in which we ''bodystormed'' \cite{klemmer2006bodies} how the ideas of dominance and submissivity could translate into specific quadruped robot behaviors. Bodystorming is a design technique in which ‘designers brainstorm by situating their bodies in the context of the interaction being designed, to gain insight into the user experience’ \cite{klemmer2006bodies}. By acting out possible interactions, designers can instinctively navigate the design space of robot behaviors and `feel' the effects of their design choices \cite{porfirio2019bodystorming}, which ultimately facilitates ideating, quickly evaluating, and iterating, as well as surfacing what often remains tacit knowledge \cite{luria2020robotic}. 

We set up the session (Figure \ref{fig:bodystorming}) in the imaginary context of a hospital corridor where a quadruped robot would be employed to carry goods and people would encounter it accidentally. This was chosen as an environment where, in the near future, accidental encounters with quadruped robots could be plausible \cite{Intel}, and also confronting, because of the physical constraints. The robot was imagined as a quadruped platform equipped with an arm. On the one hand, this was consistent with the hospital scenario in which the robot would need the arm to open doors and pick up objects; on the other hand, the arm added expressive possibilities.

In the session, which was video-recorded, the researchers enacted several walk-by situations, discussed how they felt, and shared how they interpreted each other acting. Through the discussions and the following review of the video recordings, the team distilled a set of design recommendations to inform the actual design and development of the robot behaviors, such as \emph{keeping constant speed and direction}, \emph{keeping gaze at the person}, \emph{avoiding fast changes in bodily poses}, \emph{slowing down when close to people} and more.

\subsection{Dominant and submissive behavior design}
Informed by the insights gained through the bodystorming session, we programmed two behavioral profiles for the quadruped robot \emph{Spot}, leveraging the Boston Dynamics software development kit.

The two behavioral profiles --- the dominant and the submissive (Figure \ref{teaser}) --- differ mainly in terms of gait and bodily posture. When \emph{dominant} (Figure \ref{teaser}, left), Spot trots and has a ``high height'': the legs are extended and the arm is raised. When \emph{submissive} (Figure \ref{teaser}, right), Spot crawls and has a ``low height'': the legs are bent and the arm is retracted.
Regardless of the behavioral profile, the robot's maximum speed was set as 0.5 m/s, and the obstacle padding was set as 0.5m.
A detailed overview of the behaviors can be found in the accompanying video.

\begin{figure}[ht!]
  \centering
  \includegraphics[width=1\linewidth]{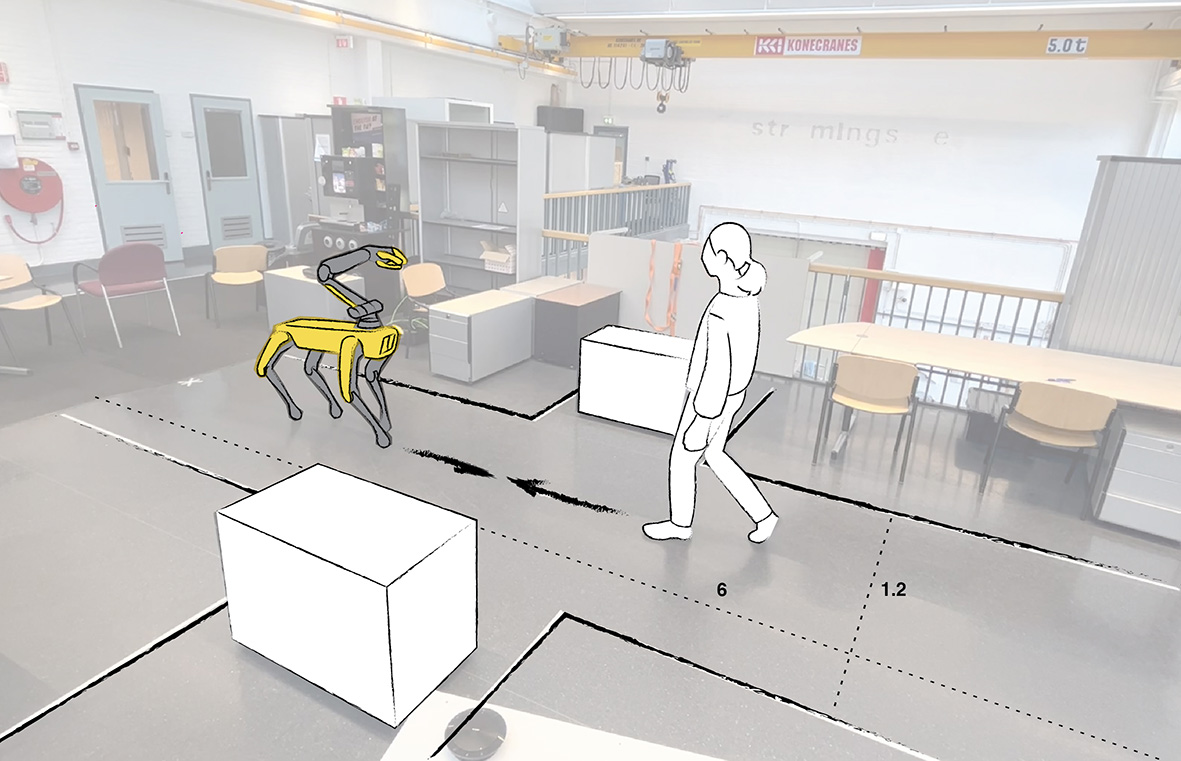}
  \caption{Head-on scenario (dominant robot depicted).}
    \label{head-on}
\end{figure}

\begin{figure}[ht!]
  \centering
  \includegraphics[width=1\linewidth]{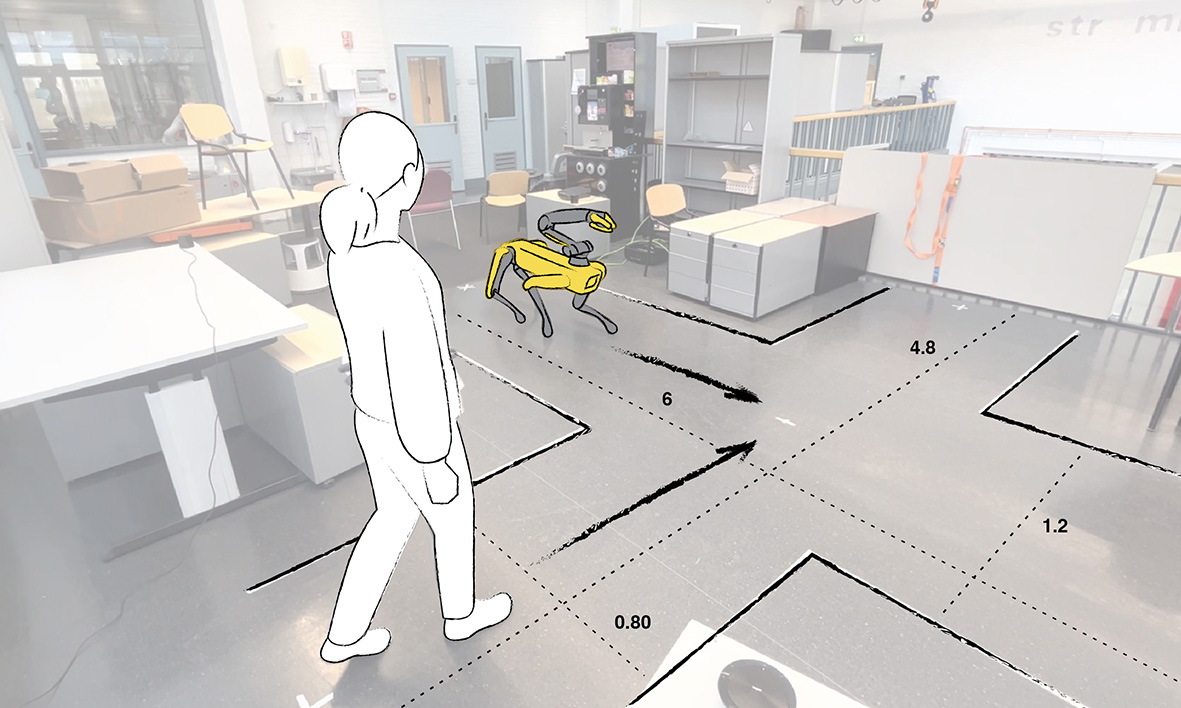}
  \caption{Crossing scenario (submissive robot depicted).}
    \label{crossing}
\end{figure}

\section{STUDY DESIGN}
We designed a 2x2 within-subjects study where we tested the two behavioral profiles (\emph{dominant} vs \emph{submissive}, Figure~\ref{teaser}) in two walking scenarios (\textit{head-on}, Figure~\ref{head-on}, vs \textit{crossing}, Figure~\ref{crossing}). 
The study protocol was reviewed and approved by the Human Research Ethics Committee of Delft University of Technology. 

\subsection{Participants}
We advertised the study through the university mailing list and a total of 21 participants volunteered to take part (13 M, 7 F, 1 non-binary; mean age $28\pm5$ years). All participants were familiar with the Spot robot: 10 participants had seen it in photos or videos before, 9 had seen it physically, and 2 had experience working with it. Hence, for our analyses, participants were categorized into two groups: with \textit{in-person} experience (n=11) and with experience based on \textit{media} (n=10). Participants expressed an overall positive attitude toward Spot: 14 participants were positive, 4 were neutral, and only 3 declared a negative attitude. 

We intentionally limited the number of participants and focused on ensuring long sessions with multiple trials (32 per person). This choice was informed by recent literature arguing for small-N study designs in psychological literature and the growing call for systematic, functional relationships as they are manifested at the individual participant level \cite{smith2018small}. 

\subsection{Setup and procedure}
The experiment was run in a lab environment, where we set up the floor either as a corridor, of 1.2 m width and 6 m length (Figure \ref{head-on}) or as an intersection, with one path of 1.2 m width and 6 m length and the other of 0.8 m width and 4.8 m length (Figure \ref{crossing}). The walkable areas were delimited by using tape and cabinets.
The two different setups were designed to make participants encounter the robot in two alternative scenarios: \textit{head-on} (the robot and the human walk on the same path, in opposite directions) and \textit{crossing} (the robot's and the human's paths intersect). Similarly to related studies \cite{hauser2023s, xu2023understanding, yang2022online}, the intersection, and the corridor especially, were chosen because they represent possible configurations of space that characterize incidental human-robot encounters \cite{hauser2023s}. These also allowed us to ``force'' participants to get close to the robot in the head-on scenario, facilitating the manifestation of desirable proxemic relations, and other behavioral responses relevant to assessing perceived safety.

Each participant experienced 16 trials of each scenario, with 8 repetitions for each condition (dominant and submissive). The order of trials was randomized within participants, and the order of scenarios was randomized across participants. Participants were not informed about the two distinct behavioral profiles beforehand.
The procedure, which lasted about 40 minutes per participant, consisted of the following steps:
\begin{itemize}
  \item \emph{Preparation}. Participants were welcomed, informed about the study procedure, and asked to fill out a written consent form, as well as a pre-experiment questionnaire.
  \item \emph{Experiment sessions}: Participants performed two blocks of 16 trials, corresponding to the two interaction scenarios. 
  For both scenarios, participants were instructed to \emph{``walk with a moderate speed towards a target marked on the floor''} and \emph{``do not collide with the robot for your safety''}. Prior to the session, participants were told to imagine a scenario where they are walking along a real corridor in a building and did not know about the presence of Spot beforehand.
\item \emph{Feedback}. Participants were invited to fill out a post-experiment questionnaire, as well as to discuss impressions and ask questions.
\end{itemize}

\subsection{Data collection}
We adopted a mixed-method approach with both behavioral and subjective data being collected and analyzed. 
Behavioral data consisted of participants’ and robot's positions at a rate of 10 Hz, which were recorded using two HTC Vive motion tracker devices (version 3.0), with four SteamVR 2.0 base stations. One motion tracker was mounted on a waist belt worn by the participants (tracker positioned on the participant's lower back), while the second tracker was mounted on top of the trunk of the robot. 

Participants' subjective data was collected through pre- and post-experiment questionnaires. The former was a general questionnaire about participants’ demographics, familiarity with quadruped robots, and general attitude towards quadruped robots. After the experiment, we asked participants to fill out the \emph{Perceived Safety Questionnaire} \cite{akalin2023taxonomy}, a 5-point semantic differential scale questionnaire that extends section V of the Godspeed questionnaire \cite{bartneck2009measurement}, and open questions about the behavioral profiles of the robot and test scenarios. We averaged the scores along the eight rated aspects of the questionnaire, to get a single perceived safety score. All questionnaires are available in the online supplementary information.

\subsection{Data analysis}
The raw trajectories of the human and the robot were used to calculate a number of metrics capturing the perceived safety. For the \textit{ head-on} scenario, for each trial, we calculated the safety margin humans adopted as the minimum distance between the human and the robot during that trial. Furthermore, to characterize how soon the participants and the robot resolved the interaction, we calculated the conflict resolution time, that is, the time it took the human and the robot to reach the minimum distance between them. For the \textit{crossing} scenario, for each time instant, we calculated the distance between the human or the robot and the center of the intersection. We then calculated the time of reaching the crossing point for the human and the robot, based on which we determined who passed the intersection first.

Participant-level quantitative data from the questionnaire was analyzed in Python using ANOVA and t-tests (as implemented in the \texttt{statsmodels} package). Trial-level trajectory metrics were analyzed in Python using logistic and linear mixed-effects models implemented in \texttt{pymer4} which included a random intercept per participant. In mixed-effects models, the submissive condition was set as a reference category for robot appearance. Open comments from participants were analyzed with inductive thematic analysis \cite{braun2012thematic}. 

\subsection{Hypotheses}
Through the combination of both behavioral and subjective data, our analyses set out to test two main hypotheses:
\begin{itemize}
    \item \emph{H1}. A quadruped robot showing submissive behaviors is perceived as safer compared to the one showing dominant behaviors. 
    \item \emph{H2}. Participants' previous in-person experience with the robot positively affects their perceived safety of the robot.
\end{itemize}

We expected to observe \emph{smaller adopted safety margin} and \emph{faster conflict resolution time} (head-on scenario) and \emph{higher likelihood of human passing the intersection first} (crossing scenario) as well as higher perceived safety ratings, when participants interacted with the Submissive robot, compared to the Dominant robot.

\section{RESULTS}
\begin{figure}[ht!]
    \centering
    \includegraphics[width=1.0\columnwidth]{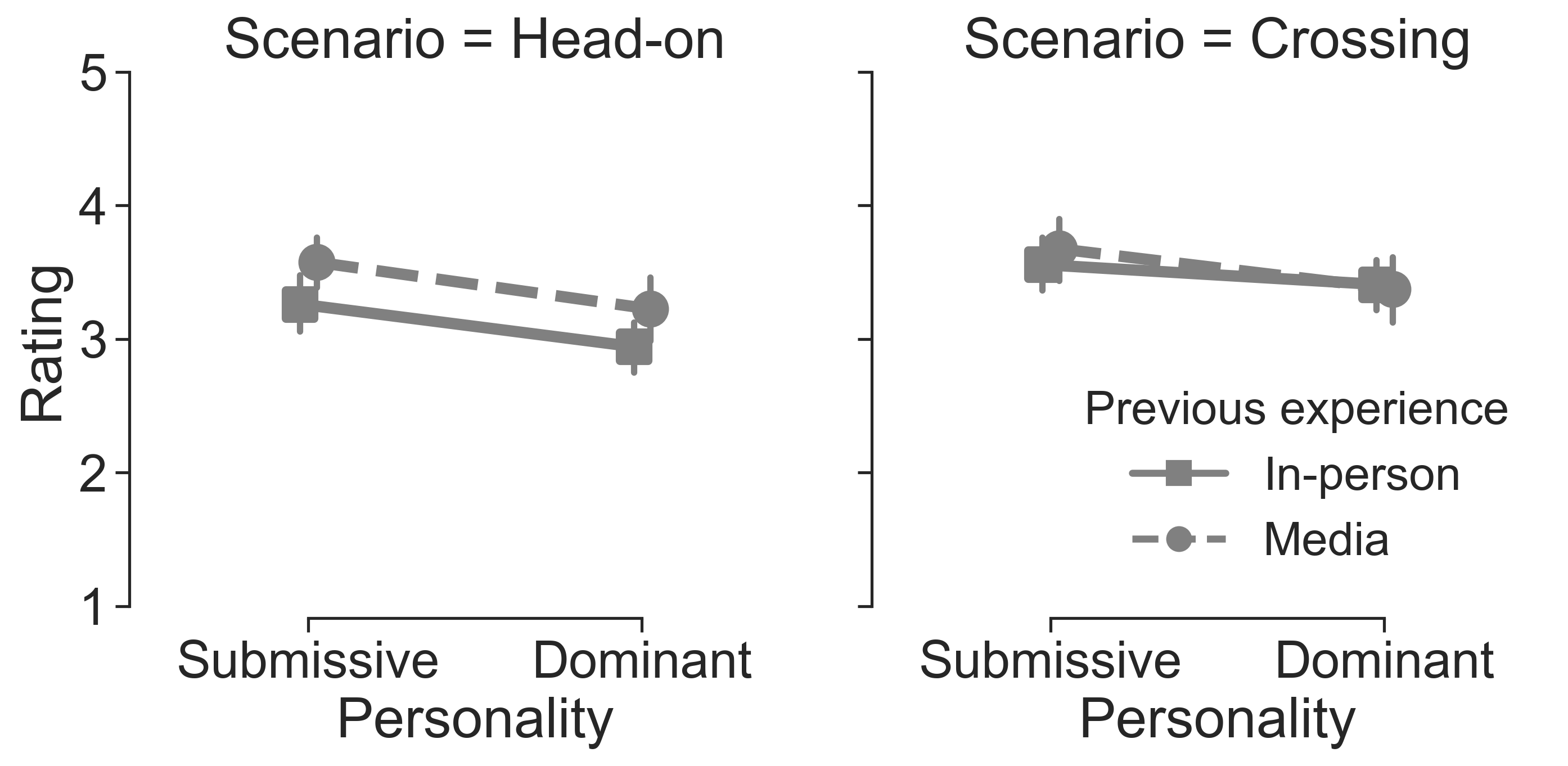}
    \caption{Average safety ratings of the robot for the two scenarios, depending on the robot's personality and participants' previous experience with the robot.}
    \label{fig:safetyratings}
\end{figure}

Participants' ratings of robot safety depended on the robot's appearance ($F=12.6, p<0.001$), participants' previous experience with the robot ($F=4.8, p=0.03$), and the interaction scenario ($F=11.1, p<0.001$) (Figure~\ref{fig:safetyratings}). There was no evidence of interactions between these three factors. Behavioral data in both scenarios did not show significant differences between conditions and participant groups.

\subsection{Subjective perception of safety is higher when the quadruped robot is submissive}

\subsubsection{Subjective perception}
Participants rated interactions with the submissive robot higher in terms of safety, as compared to the dominant robot (3.5 vs 3.2, $t=4.5$, $p<0.001$). The ratings reported for the head-on scenario were consistently lower than for the crossing scenario (3.2 vs 3.5, $t=-5.8$, $p<0.001$). 

The responses to the open questions further emphasized that many participants (11/21) had a more negative impression of the dominant robot, as compared to the submissive robot. Participants described the dominant robot as being more \emph{intimidating} (P2, P7, P12, P19), \emph{scary} (P6, P13, P14) and \emph{threatening} (P18), but also \emph{rude} (P17) and \emph{arrogant} (P5). While some participants also described it as more alert (P10), strong (P12), and powerful (P6, P9), these attributes were generally expressed in negative terms. As P10 further explained: \emph{"I felt more need to walk sideways. Especially when the Spot was at a higher height, I was worried I’d touch it when swinging my arms if I just walked turned completely forward"}. Similarly, P4 mentioned: \emph{"Spot never felt like passing safe when in high eight stance"}. Some further explained that gaze was an important aspect affecting their negative perception (P12, P14, P17). P12 in particular mentioned: \emph{"when the spot was with a higher height it matched with my eye level and this made me feel a bit intimidated"}.

On the contrary, some participants (P4, P5, P7, P14, P15) explicitly reported positive remarks regarding the submissive robot. It was described as being \emph{less threatening} (P14) more \emph{predictable} (P14) and \emph{stable} (P15), but also more \emph{cooperative} (P4) and \emph{cautious} (P7). As P7 further explained: \emph{"Lower height Spot is more cautious and slower, therefore, I assume it will detect me on time and no sudden surprise"}.

Only three participants  (P7, P10, P17) described the dominant robot in less negative terms, compared to the submissive one. Remarkably, all of these participants explicitly made references to animals for reasoning about the robot. The only real exception in terms of attitude, however, was P17, the only one who reported a positive view of the dominant Spot and a negative on the submissive. As they explained: \emph{"The lower posture was canine, so felt both familiar but potentially threatening. Also, it’s the posture of an animal in stalking, so felt like it might pounce… The more canine crouch I assumed was in its own world and didn’t expect it to move for me"}. 

\begin{figure}[ht!]
  \centering
  \includegraphics[width=1\linewidth]{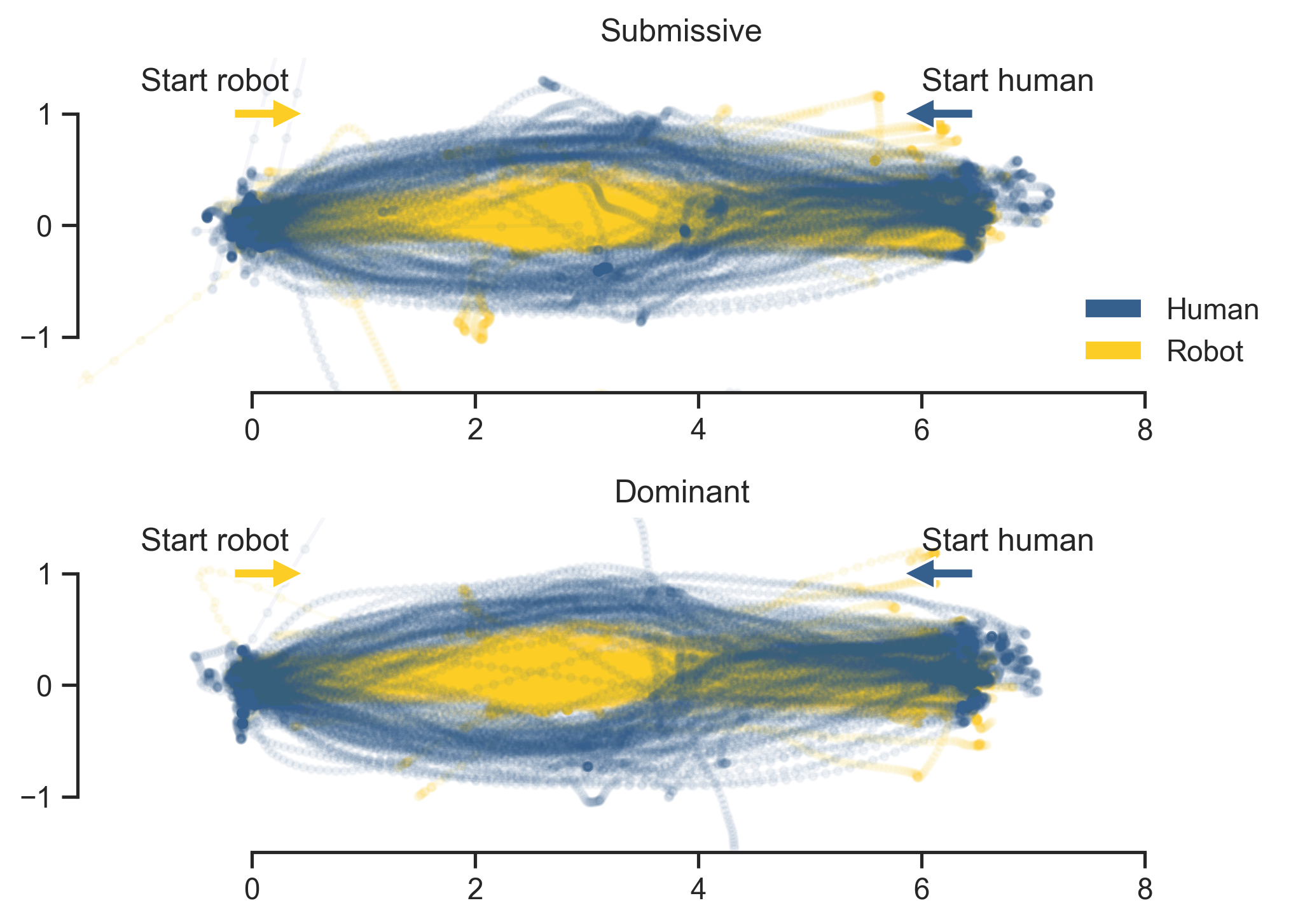}
  \caption{Head-on scenario: Trajectories of all human participants and the robot; each line represents one trial. Arrows represent walking directions of the human and the robot.}
      \label{fig:head_on_trajectories}
\end{figure}

\subsubsection{Behavioral response}
In the head-on interactions, the participants typically walked around the robot; the robot deviated from the straight-line path to its goal to a lesser extent than the participants (Figure~\ref{fig:head_on_trajectories}).

To further investigate interaction dynamics in the head-on scenario, we analyzed the safety margin and the conflict resolution time. We found no evidence of a difference between the dominant and submissive conditions in terms of both the safety margin ($b=0.008, t=0.7, p=0.5$) and the conflict resolution time ($b=0.22, t=1.8, p=0.07$).

\begin{figure}[ht]
  \centering
  \includegraphics[width=1\linewidth]{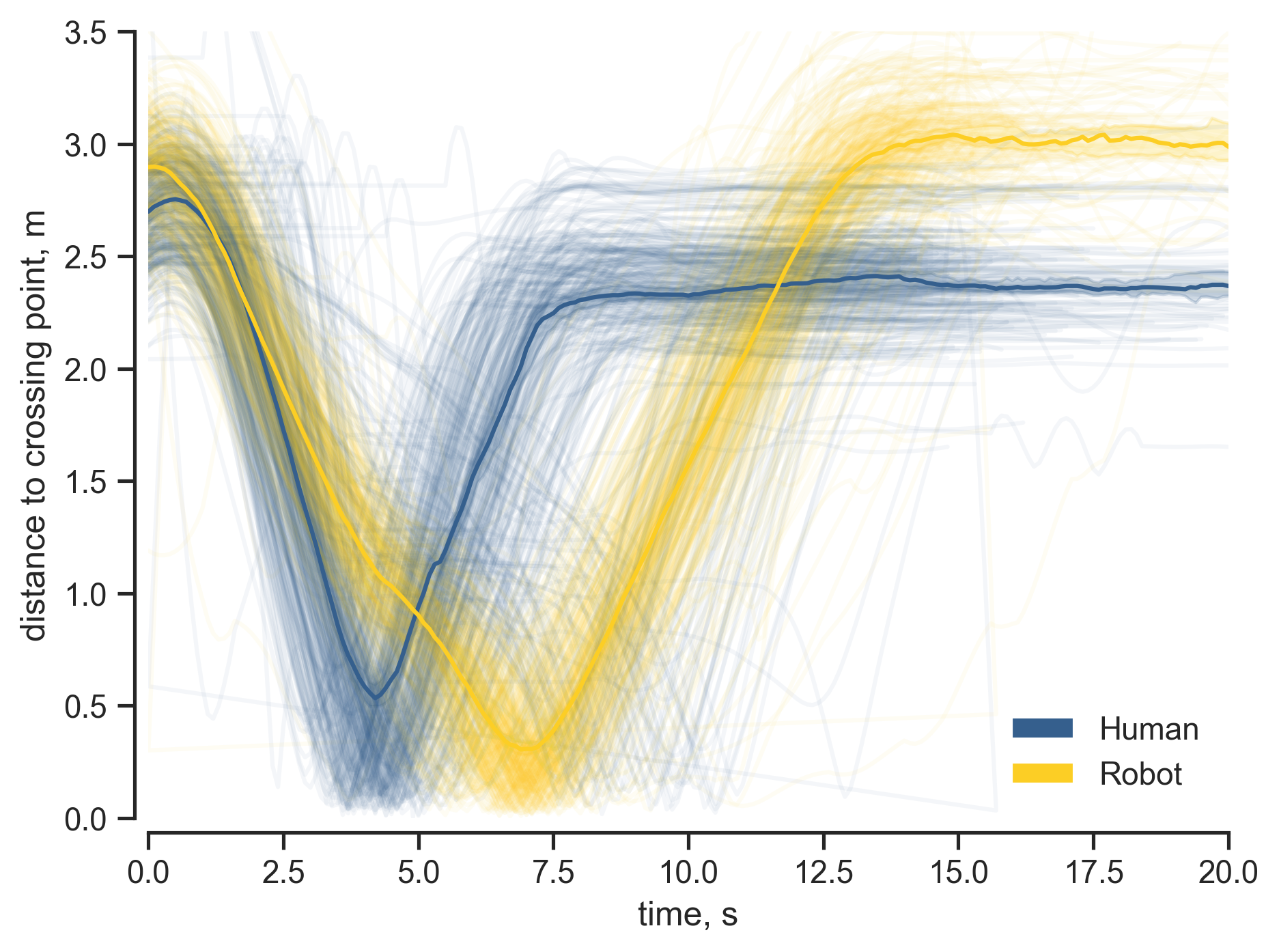}
  \caption{Crossing scenario: Distance to the crossing point for the human participants and the robot. Faint lines represent individual trials, thick lines represent the medians.}
  \label{fig:crossing_distance}
\end{figure}

In the crossing scenario, participants predominantly reached the crossing point before the robot (Figure~\ref{fig:crossing_distance}): that was the case in $81.5\%$ of all crossing trials, while the robot reached it first in $18.5\%$ of trials. We did not find evidence for a relationship between the outcome of the crossing interaction and the appearance of the robot ($b=-0.13, z=-0.4, p=0.7$) or participants' previous experience ($b=1.1, z=0.97, p=0.3$).

\subsection{Previous in-person experience has a small negative effect on perceived safety}
\subsubsection{Subjective perception}
Participants with previous in-person experience rated the robot's safety lower (Figure \ref{fig:safetyratings}) than participants who were only familiar with the robot based on media (online videos) (3.3 vs 3.5, $t=-2.1$, $p=0.03$). 

This result contrasts with our assumption that knowledge only based on media would encourage a more negative perception of the robot. While some participants explained that their perception of the robot improved after some trials (P5, P9, P19), others mentioned how the very interaction with the robot lowered their confidence around it. In particular, the way the robot moved from one appearance to another, or from one direction to the opposite, was perceived as threatening. For instance, P17 mentioned: \emph{"My perception was different when it lowered or raised its head at the last moment, it felt more threatening..."}. Some participants further explained how sudden and inconsistent movements represented sources of unpredictability, that undermined the perceived safety of the robot. E.g., P2 mentioned: \emph{"Spot’s behavior when turning around affected how confident I felt. For example, when it seemed to stumble I felt less comfortable being close to it afterward"}. Similarly, P15 explained: \emph{"I saw the robot reacted differently, i.e. sometimes it kept walking. sometimes it stops. such inconsistency may scare some people as the robot's motion seems unpredictable"}.

\subsubsection{Behavioral response}
In the head-on scenario, we found no evidence of a relationship between participants' previous experience with the robot and safety margin ($b=-0.001, t=-0.03, p=0.97$) or conflict resolution time ($b=0.21, t=0.5, p=0.6$).

\section{DISCUSSION AND CONCLUSIONS}
Our results highlighted that participants perceived the submissive robot as safer compared to the dominant one, whereas the behavioral dynamics of interactions did not change depending on the robot's appearance. In head-on interactions with both submissive and dominant robots, humans---rather than the robot--- resolved the interactions. Participants' previous in-person experience with the robot was associated with lower subjective safety ratings, but, similarly to robot appearance, did not correlate with the interaction dynamics.

Our findings further contribute to existing knowledge by providing novel insights into \emph{appearance-constrained robot communication} and its implications, an area that today remains under-investigated \cite{sanoubari2022message}. We provide further nuances on the effects that specific features of the robot behaviors (posture, gaze) have on the safety and comfort perceived by the human when walking in a constrained space with a co-present robot. Xu et al. \cite{xu2023understanding}, in particular, have previously found that a robot in a facing direction but without gazing at participants can make people feel more secure and less uncertain about the robot's possible behaviors. Our study evidenced a similar effect. People consider Spot as more threatening when in the dominant version, as it would be perceived to be looking at people in the eye. Interestingly, the robot has no human/zoomorphic capacity to see, as what people refer to as eyes and face are actually the hand, yet these have a strong impact on perceived safety and comfort.

Further, we learned about the importance of understanding \textit{individual imaginaries} (the reminiscence of the hunting dog) as a key to understanding possible variations in people's perceptions of robots. The principle is not novel per se, but the way this materialized in our study is relatively unprecedented. Specifically, what we learned regarding the effect of animal reminiscence– canine in particular– adds to existing knowledge regarding the influence that people’s familiarity with pets can have on HRI. Takayama and Pantofaru \cite{takayama2009influences} previously found that people with pet ownership experience would distance themselves closer to robots. Yet our work showed that certain \emph{bodily poses may recall specific animal behaviors with a significant impact on human perception and attitude}, e.g., a crouching quadruped robot may be expected to suddenly jump or approach if framed as a hunting dog. This phenomenon may be so impactful that it could generate an opposite perception of the robot.

\subsection*{Limitations}
Although insightful, the results did not completely align with our hypotheses. This may be partially due to a limitation of our study: the lack of diversity in the participants' sample. Our participants were predominantly young, male, highly educated, and overall familiar with quadruped robots, in line with the field's tendency to involve people from Western, educated, industrialized, rich, and democratic societies \cite{seaborn2023not}. Thereafter, our findings can't be generalized for all possible populations that quadruped robots may encounter if were to be employed in civil environments. We especially lack data about the social perception of the robot from the perspective of people who didn't have any knowledge about it before and, even more so, who might have low technological fluency. Thus, future research should focus on understanding whether aspects of technological fluency, age, gender, or ethnicity might relate to different perceptual responses toward the quadruped robot and its bodily expressivity.

Nevertheless, we believe this work brings a clear contribution to the HRI field not solely because it provides insights into aspects hardly addressed before --the effect of bodily expressivity on the perceived safety of quadruped robots-- but also, and foremost, because through this work we invite the field to reflect on the epistemic assumptions we make when studying social robot perception (and the limits of quantitative measures) and argue for the importance of understanding felt experiences.

\section*{Data Availability}
Online supplementary information (including all the code and data generated in this study) is available at \href{https://osf.io/zd2fa}{https://osf.io/zd2fa}

\section*{ACKNOWLEDGMENT}
We would like to thank Kseniia Khomenko for the technical support with the Spot robot.

\bibliographystyle{IEEEtran}
\bibliography{references}

\end{document}